\documentclass[twoside]{article}

\usepackage[accepted]{aistats2023}
\usepackage[round]{natbib}

\usepackage{graphicx}
\usepackage{amsmath}
\usepackage{amsfonts}
\usepackage{xcolor,colortbl}
\usepackage{mathtools}
\usepackage{url}
\usepackage{algorithm2e}
\usepackage{booktabs, multirow} 
\usepackage{soul}
\usepackage{subcaption}
\usepackage{spverbatim}

\newcommand{\bs}[1]{\boldsymbol{\mathbf{#1}}}

\usepackage{booktabs,subcaption,amsfonts,dcolumn}

\usepackage{tikz}
\usepackage{float}
\usetikzlibrary{fit,positioning,matrix}
\usetikzlibrary{matrix,arrows,decorations.pathmorphing}
\tikzset{
    node style  one/.style={draw,fill=black!60, circle,minimum size=\myunit},
    node style  zero/.style={draw,circle,minimum size=\myunit},
    node style  blue/.style={draw,fill=blue!60, circle,minimum size=\myunit},
    node style  red/.style={draw,fill=red!60, circle,minimum size=\myunit},
    node style  green/.style={draw,fill=green!60, circle,minimum size=\myunit},
    node style  na/.style={ draw,fill=white, circle,dashed, inner sep=1.3mm,minimum size=\myunit},
    node style    sp/.style={draw,circle,minimum size=\myunit},
    node style    ge/.style={circle,minimum size=\myunit},
    arrow style  mul/.style={draw,sloped,midway,fill=white},
    arrow style plus/.style={midway,sloped,fill=white}
}

\begin{document}

%

%

\twocolumn[

\aistatstitle{Learning from Multiple Sources for Data-to-Text and Text-to-Data
}



\aistatsauthor{ Song Duong\textsuperscript{\textnormal{1,2}} \And Alberto Lumbreras\textsuperscript{\textnormal{1}} \And  Mike Gartrell\textsuperscript{\textnormal{1}} \And Patrick Gallinari\textsuperscript{\textnormal{1,2}}}
\vspace{2mm}
\aistatsaddress{ \textsuperscript{\textnormal{1}}Criteo AI Lab\\\textsuperscript{\textnormal{2}}Sorbonne Université, CNRS, ISIR,  Paris, France}]
\runningauthor{Duong, Lumbreras, Gartrell, Gallinari}


\begin{abstract}
  Data-to-text (D2T) and text-to-data (T2D) are dual tasks that convert structured data, such as graphs or tables into fluent text, and vice versa. These tasks are usually handled separately and use corpora extracted from a single source. Current systems leverage pre-trained language models fine-tuned on D2T or T2D tasks. This approach has two main limitations: first, a separate system has to be tuned for each task and source; second, learning is limited by the scarcity of available corpora. This paper considers a more general scenario where data are available from multiple heterogeneous sources. Each source, with its specific data format and semantic domain, provides a non-parallel corpus of text and structured data. We introduce a variational auto-encoder model with disentangled style and content variables that allows us to represent the diversity that stems from multiple sources of text and data. Our model is designed to handle the tasks of D2T and T2D jointly. We evaluate our model on several datasets, and show that by learning from multiple sources, our model closes the performance gap with its supervised single-source counterpart and outperforms it in some cases.
  \footnote{The code is available at \url{github.com/sngdng/MSUnsupVAE}.}
\end{abstract}

\section{INTRODUCTION}
\label{sec:introduction}


\begin{figure}
    \centering
    \includegraphics[width=\linewidth]{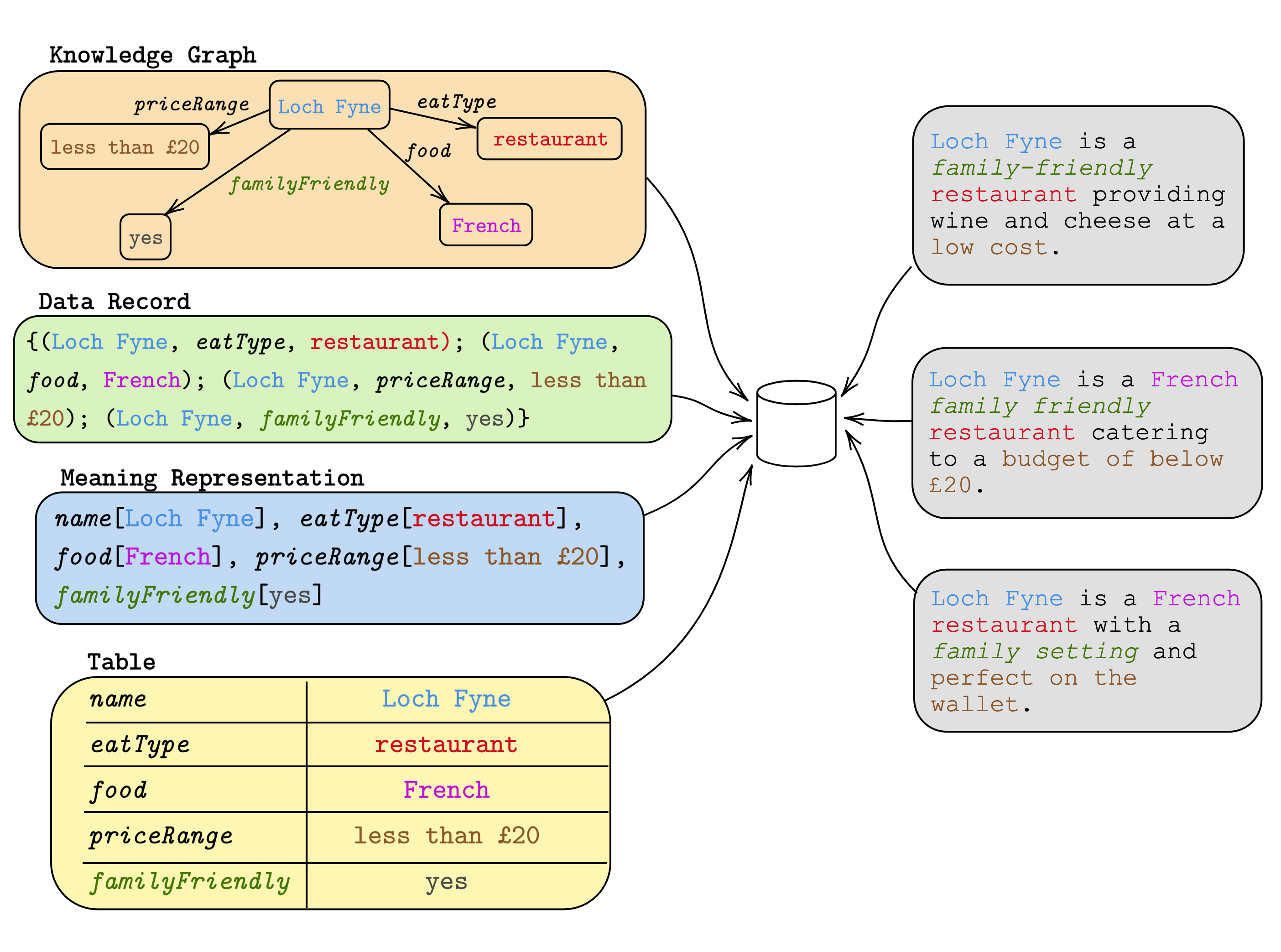}
    \caption{The same content can be represented by different structures and by different texts. We show here an example from the E2E dataset \citep{e2e_cleaned}. On the left side, from top to bottom: knowledge graph, data record, meaning representation, and table.}
    \label{fig:many2many}
    \vspace{-8pt}
\end{figure}

Data-to-text (D2T) \citep{kukich-1983-design, mckeown_1985} is the task of converting structured data into fluent text. This generic task is involved in several applications: generating summaries from patient records \citep{Scott2013DatatotextSO} or from tables such as summaries of sports games \citep{wiseman-etal-2017-challenges}, generating product descriptions from databases in e-commerce \citep{wang-etal-2017-statistical}, and so on. What makes this task different from conventional text generation is that structured data comes in many forms, e.g. web tables, databases, knowledge graphs, and so D2T systems have to handle these heterogeneous formats. Text vocabulary can also vary greatly from one domain to another, and systems need to adapt to this diversity as well (see Fig.~\ref{fig:many2many}). In addition, both modalities, text and data, are often noisy and incomplete, and in most practical situations there is no aligned corpus available, or at best small hand-crafted corpora. As a concrete example, consider the generation of product descriptions on an e-commerce website. Items can be described either as structured tables extracted from databases or as text. The former provides details of the characteristics of the items, while the latter summarises this information for the user. Information can be obtained from different providers and in different formats, also known as different sources \citep{heterogeneous}. Even from the same vendor, descriptions can vary widely depending on the product category (e.g. high-tech products and clothing). They may be quite noisy, such as unstructured sentences or series of keywords instead of a plain description, and in many cases only the text or data is available. In this case, the supervised formulation falls short, as we need to be able to use non-parallel data.

Text-to-data (T2D) is the dual task that aims at extracting relevant information from text and identifying their relations in order to generate structured data in a pre-defined format. Current T2D approaches are mainly limited to the conversion from text to graphs \citep{ Ye_Zhang_Deng_Chen_Tan_Huang_Chen_2021, agarwal-etal-2020-machine, paolini_tanl}. Until now, both tasks have been handled separately through different methods.

Recent advancements in LLMs have enabled encoding various forms of information in text and viewing D2T and T2D as dual tasks within a unified framework. However, existing systems have primarily focused on simplified scenarios to meet these requirements:

(i) First, to the best of our knowledge, all models to date consider a single source for data and text, and deal with a data format and a text style as defined by the training corpus. However, in practice, one is faced with heterogeneous formats, where data is collected from multiple sources, each with its own specificity. 
This also holds for related NLP tasks including \textit{Question Answering} \citep{chen-etal-2020-hybridqa, talmor-berant-2019-multiqa, nan-etal-2022-fetaqa} and \textit{Formal-Language-to-Text }\citep{shu-etal-2021-logic, chen-etal-2020-logic2text}. 
This variety of data sources has led to the development of systems adapted to each case, resulting in architectures designed specifically for knowledge graphs \citep{ke-etal-2021-jointgt, zhao-etal-2020-bridging}, tables \citep{liu2017table,rebuffel:hal-03479813, 9026768, Liu_Luo_Xia_Ma_Chang_Sui_2019}, and so on. Recent work unifies multiple sources of structured data while considering a diversity of tasks, such as \textit{Structured Knowledge Grounding} (SKG) \citep{UnifiedSKG}, by grounding both user requests and structured data in a common structured template. 

(ii) Secondly, most models are trained in a supervised setting that requires the manual construction of aligned datasets. The latter are usually limited in size; for instance, one of the main D2T reference datasets, WebNLG \citep{gardent2017creating}, is composed of only 35,000 aligned pairs of text and knowledge graphs. Some recent works \citep{jin-etal-2020-genwiki, lebret-etal-2016-neural, chen-etal-2021-wikitablet} have proposed larger datasets, e.g., by collecting texts from Wikipedia and extracting knowledge graphs from DBPedia \citep{dbpedia} or Wikidata \citep{wikidata}. However, as these text-graph pairs have been generated automatically, they are usually loosely aligned or simply non-parallel. Some works \citep{schmitt-etal-2020-unsupervised, guo-etal-2020-cyclegt, DBLP:conf/aistats/GuoJ0Q0ZZW21} have tackled the non-parallel setting in the presence of a single source of knowledge graphs.

This paper considers the dual tasks of D2T and T2D within a unified formalism for multi-source conversion. We assume that structured data and texts are collected from multiple sources, with each source in a specific structured format. Our objective is to develop a unique system capable of performing conversion in both directions for a specified number of sources, where we convert data from a given source to text and, given a text, we convert it to any format from the set of targeted sources. With parameters shared between sources and tasks, the model can take full advantage of all sources, as opposed to systems trained on each source independently. We make the additional assumption that for training, data are collected from each source independently and are not aligned, i.e., there is no unique correspondence between a text and data. Given the difficulty of generating parallel datasets for D2T and T2D, this is a realistic scenario that allows us to leverage larger corpora in many practical situations. Note that even for aligned corpora, most of them are generated automatically, so the correspondence can be extremely loose. Compared to the usual single-source supervised or unsupervised settings, where a one-to-one mapping is assumed between text and data, we consider here a new problem corresponding to a many-to-many mapping between structured data and text: data comes from multiple heterogeneous sources, while text can express the same content in multiple ways.

Our contributions are the following:

$\bullet$ We frame D2T and T2D as two complementary sequence-to-sequence tasks that can be alternately optimized in an end-to-end, cycle-training fashion via iterative back-translation.

$\bullet$ We propose a latent variable model for handling multi-source D2T and T2D through a shared model common to both tasks and to all the sources. Our approach is based on a pre-trained T5 language model \citep{T5_JMLR:v21:20-074} and leverages the variational autoencoder (VAE) formulation.

$\bullet$ We evaluate our method on several D2T datasets. We show that by learning from multiple sources, our model closes the performance gap with its supervised single-source counterparts on many datasets and even outperforms them on the DART dataset \citep{nan-etal-2021-dart}. This demonstrates the utility of learning from multiple heterogeneous sources in a unified way. Although, for practical reasons, our experiments have been conducted with the ``T5-small'' model, the results remain competitive with current state-of-the-art obtained with larger models on a portion of the datasets.



\section{RELATED WORK}
\label{sec:relatedworks}
\textbf{Data-to-text (T2D).} Recent neural approaches have trained end-to-end models with either encoder/decoder architectures \citep{wiseman-etal-2017-challenges, gardent-etal-2017-webnlg}, or using neural networks as part of a pipeline system \citep{moryossef-etal-2019-step, castro-ferreira-etal-2019-neural, Puduppully_Dong_Lapata_2019, laetal}. In particular, using large pre-trained language models has been shown to greatly improve performance, either by simply framing data-to-text as a text-to-text task \citep{kale-rastogi-2020-text}, or by using a GNN-based planner for graph-to-text tasks \citep{ribeiro-etal-2021-investigating, zhao-etal-2020-bridging, guo-etal-2020-p2}. 
Current works mainly consider single-source problems. Some works have considered leveraging multiple sources, either by aggregating multiple structured datasets into a single-format dataset \citep{chen-etal-2020-kgpt,nan-etal-2021-dart}, or by grounding user requests and structured data in a common structured template \citep{UnifiedSKG}.

\textbf{Text-to-data (T2D).} Converting unstructured text to structured data can be framed as a specific instance of (open) information extraction tasks, such as Named Entity Recognition (NER) \citep{nadeau2007survey, putthividhya-hu-2011-bootstrapped} or, more commonly, joint entity and relation extraction. Generally, given a text, the T2D task requires identifying the relevant entities and the relations between them, with the result being encoded as a triplet (Entity1, Relation, Entity2). Recent approaches consider the graph as a sequence of triples and cast the problem as a text-to-text task. They typically combine an encoder with an auto-regressive decoder \citep{ Ye_Zhang_Deng_Chen_Tan_Huang_Chen_2021, agarwal-etal-2020-machine, paolini_tanl}. There has been less work on converting text to tables, although some recent approaches aim to generate tables from queries \citep{Gupta-Jigsaw} in the information retrieval domain.


\textbf{Cycle training with back-translation for unsupervised learning.}
Back-translation \citep{cheng-etal-2016-semi, NIPS2016_He_bt, cotterell+kreutzer.arxiv18} has been shown to be very effective in unsupervised machine translation \citep{artetxe2018unsupervised, lample2018unsupervised}. Given non-parallel samples $(x,y)$ from a source and target language, this approach jointly learns two models (from one language to another and vice-versa) by leveraging cycle-consistencies $x \rightarrow \hat{y} \rightarrow x$ and $y \rightarrow \hat{x} \rightarrow y$. \cite{lample2018unsupervised} identified a fundamental principle for successful back-translation, namely that each model must be trained on a specific ``language modeling objective to generate correct sentences in its language'', i.e., each model must be able to capture the specifics of its own source domain.

Because of the duality of D2T and T2D, some works have framed both tasks as translation tasks between languages by linearizing the structured data. Following \cite{lample2018unsupervised}, \cite{schmitt-etal-2020-unsupervised} introduced a denoising objective for graph-to-text, and apply iterative back-translation to train a single sequence-to-sequence model (LSTM with attention and a copy mechanism) to perform both graph-to-text and text-to-graph. \cite{guo-etal-2020-cyclegt} also leveraged back-translation to perform unsupervised learning, but considered separate graph-to-text and text-to-graph components. The former component is based on a pre-trained language model and takes a linearized graph as input, while the latter component uses an off-the-shelf entity extraction model and simply learns a relation classification model to obtain the graph. This breaks the symmetry of the two tasks, and it seems that the denoising objectives of \cite{schmitt-etal-2020-unsupervised} are no longer needed for back-translation to succeed. A few recent works have incorporated cycle-consistent approaches for the semi-supervised setting. \cite{domingo2022multitask} propose training a single T5 model for both graph-to-text and text-to-graph tasks, while \cite{chang-etal-2021-neural} proposes various data augmentation approaches and uses cycle consistency to ensure that the augmented data samples can be correctly reconstructed after having been formulated as text, and vice versa.

\textbf{Variational framework for data-to-text.}
The variational autoencoder (VAE) \citep{Kingma2014, DBLP:conf/icml/RezendeMW14} is a generative model capable of leveraging latent variables. \cite{bowman-etal-2016-generating} was the first to train a text VAE using an RNN encoder/decoder model, with KL annealing used to mitigate posterior collapse. 

In the D2T literature, variational models are used to induce diversity or to enable more fine-grained control over text generation from data. \cite{DBLP:conf/aistats/GuoJ0Q0ZZW21} proposed a latent variable model to represent the diversity in text, allowing the model to learn a one-to-many mapping from the graph to the text domain. This approach performs both graph-to-text and text-to-graph tasks and uses the same standard NER component as in \cite{guo-etal-2020-cyclegt}. \cite{Ye2020Variational} proposed a variational model with disentangled content and style latent variables to perform the graph-to-text task. Their semi-supervised algorithm is also inspired by back-translation but still requires some parallel data. For the table-to-text task, \cite{wiseman-etal-2018-learning} modeled templates as a sequence of discrete latent variables and learns them using hidden semi-Markov models. \cite{puduppully-etal-2022-data-variational} proposed a variational approach to model latent plans in order to perform content planning in D2T generation. \cite{chen-etal-2021-deconfounded} introduced a VAE to model the causality of logical table-to-text as a latent variable.

\section{PROBLEM DEFINITION}
\label{sec:unsuplearning}

\textbf{Multi-source scenario.} We consider the scenario where we have multiple data sources at our disposal, and the data from each source are collected independently. Each source consists of a set of texts $\{x_n\}_{n=1}^{N}$ and a set of single-format structured data $\{y_m\}_{m=1}^{M}$. 
The formats of the different sources may be heterogeneous, e.g., tables of different types, graphs, etc.  
We consider structured data formats that can be represented or mapped to a set of relational triples (subject, relation, object), as is the case for graphs, data records, and tables (see Fig.~\ref{fig:many2many}).

\textbf{D2T \& T2D tasks.} Our goal is to solve  the D2T and T2D tasks.  For the D2T task, we are given structured data from a source, and we seek to generate fluent text that explains the data. Conversely, for the T2D task, we are given a text and a target format indicator, and we aim to generate the corresponding structured data that summarize the text. We make the assumption that for D2T, multiple sentences can describe the same data, since there are many ways to verbalize a given piece of content. For T2D, given a target format, we assume that the mapping is unique since the content is uniquely encoded in the text.

\textbf{Non-parallel setting.} We study a general setting where each source consists of non-parallel sets of text and data, respectively: $(X_N, Y_M) = \left(\{x_n\}_{n=1}^{N},\{y_m\}_{m=1}^{M}\right)$. In other words, during training, we assume that the origin of the text and data is known, but their alignment is unknown. Our goal is to train a single system that can perform the conversion in both directions for all sources. 
We denote $p_{\mathrm{src}}(x,y)$ as the unknown joint distribution of the text and data from a source. In a non-parallel setting, since we do not know the correspondence, the samples in the dataset $(X_N, Y_M)$ are assumed to come from the marginals, which are respectively $p_T$ and $p_D$ for the text and the data:
\begin{align}
    p_T(x)&= \int_{y}p_{\mathrm{src}}(x, y)dy, \\
    p_D(y) &=\int_{x}p_{\mathrm{src}}(x, y)dx.\label{eq:hyp2}
\end{align}
We assume that text and data are not statistically independent, i.e., \ $p_{\mathrm{src}}(x,y) \neq p_T(x)p_D(y)$, so that it is possible to learn mappings $p_T(x|y)$ and $p_D(y|x)$ between the text and the data.

\section{BACKGROUND: ITERATIVE BACK-TRANSLATION}
\label{sec:ibt}

Before delving into our proposed model in Section \ref{sec:augmented-model}, in this section we formalize iterative back-translation, which is a fundamental part of our model derivation, and formally link our approach to the classical back-translation loss objective (Eq.~\ref{eq:cycle_obj}). In the discussion that follows, we consider for simplicity a single-source objective. This is not limiting, since the multi-source problem amounts to solving several single-source problems with a shared model.

\begin{figure}
\centering
\resizebox{0.45\columnwidth}{!}{\begin{tikzpicture}
\newcommand{\mynodedistance}{5mm}
\newcommand{\mynodesize}{10mm}
\tikzstyle{observed}=[circle, minimum size = \mynodesize, thick, draw =black!80, fill = black!10, node distance = \mynodedistance]
\tikzstyle{latent}=[circle, minimum size = \mynodesize, thick, draw =black!80, node distance = \mynodedistance]
\tikzstyle{latent_invisible}=[circle, minimum size = \mynodesize,  node distance = \mynodedistance]
\tikzstyle{parameter}=[circle, minimum size = \mynodedistance, node distance = \mynodedistance]
\tikzstyle{connect}=[-latex, thick]
\tikzstyle{plate}=[rectangle, draw=black!100, rounded corners=0.25cm]
\tikzstyle{plate_invisible}=[rectangle, rounded corners=0.25cm]
  \node[observed] (x_t) {$x_n^{(t)}$};
  \node[latent]   (y_t) [right=2cm of x_t]{$y_n^{(t)}$};
  \node[parameter] (theta_t) [below =0.8cm of x_t]{$\theta^{(t)}$};
  \path (y_t) edge [connect] (x_t);
  \node[plate, 	 fit= (x_t)(y_t), inner sep=4.6mm, xshift=-1mm]() {}; 
  \node[rectangle, fit= (y_t), thick, inner sep=0mm, label=below right:$N$, xshift=-3mm, yshift=0.5mm] {};

  \node[latent] (x_d) [below=2.5cm of x_t] {$x_m^{(d)}$};
  \node[observed]   (y_d) [below=2.5cm of y_t]{$y_m^{(d)}$};
  \node[parameter] (theta_d) [below=0.8cm of y_t]{$\theta^{(d)}$};
  \path (theta_d) edge [connect] (y_d);
  \path (theta_d) edge [connect] (y_t);
  \path (theta_t) edge [connect] (x_t);
  \path (theta_t) edge [connect] (x_d);
  \path (x_d) edge [connect] (y_d);
  \node[plate, 	 fit= (x_d)(y_d), inner sep=4.6mm, xshift=-1mm]() {}; 
  \node[rectangle, fit= (y_d), thick, inner sep=0mm, label=below right:$M$, xshift=-3mm, yshift=0.5mm] {};
\end{tikzpicture}}
\caption{Graphical representation of the unsupervised setting where aligned pairs are not available.}
\label{fig:nonaugpgm}
\end{figure}
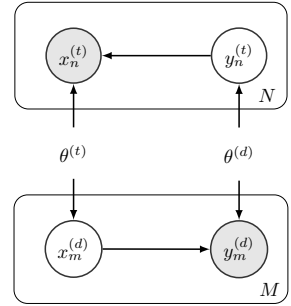

In the non-parallel setting, we do not have access to ground-truth parallel pairs $(x,y) \sim p_{\mathrm{src}}(x,y)$, which makes the learning of $p_T(x | y)$ and $p_D(y | x)$ difficult since we cannot directly maximize the conditional likelihoods. We address this issue by introducing latent variables $y_n^{(t)}$ and $x_m^{(d)}$, corresponding respectively to the unobserved ground-truth pair or \textit{missing pair} for our observed text $x_n^{(t)}$ and data $y_m^{(d)}$, as shown in Fig~\ref{fig:nonaugpgm}.
Our goal is to learn the densities $p_T(x | y)$ and $p_D(y | x)$, which we parameterize respectively with $\theta^{(t)}$ and $\theta^{(d)}$. We denote by $\theta = [\theta^{(t)},\theta^{(d)}]$ the set of parameters of our model.
Given any arbitrary observed text $x_n^{(t)}$, we can express its log-likelihood using the missing data distribution $p_{\theta^{(d)}}(y_n^{(t)})$ (see Appendix~\ref{sec:loglikelihood_latent_variable_model}) as
\begin{align}
\log p_{\theta}(x_n^{(t)}) 
&= \underset{p_{\theta^{(d)}}(y_n^{(t)} )}{\mathbb{E}} [ \log p_{\theta^{(t)}}(x_n^{(t)} | y_n^{(t)})] \notag\\
&+ D_{KL}(p_{\theta^{(d)}}(y_n^{(t)} ) \| p_{\theta^{(d)}}(y_n^{(t)} | x_n^{(t)})).
 \label{eq:loglikelihood}
\end{align}
Likewise, the log-likelihood of any given data $y_m^{(d)}$ can be obtained the same way using $p_{\theta^{(t)}}(x_m^{(d)})$:
\begin{align}
\log p_{\theta}(y_m^{(d)})&=\underset{p_{\theta^{(t)}}(x_m^{(d)} )}{\mathbb{E}} [ \log p_{\theta^{(d)}}(y_m^{(d)} | x_m^{(d)})] \notag\\
&+ D_{KL}(p_{\theta^{(t)}}(x_m^{(d)} ) \| p_{\theta^{(t)}}(x_m^{(d)} | y_m^{(d)})). \label{eq:loglikelihood2}
\end{align}
Given each equation, we do not know the distribution of the missing data $p_{\theta^{(d)}}(y_n^{(t)})$  nor of the missing text $p_{\theta^{(t)}}(x_m^{(d)})$. An arbitrary choice of these distributions would lead to an arbitrary estimation of the mapping functions $p_{\theta^{(t)}}(x_n^{(t)} | y_n^{(t)})$ and $p_{\theta^{(d)}}(y_m^{(d)} | x_m^{(d)})$. Instead, we follow the iterative back-translation scheme proposed in \cite{hoang-etal-2018-iterative} and consider that
\begin{align}
     p_{\theta^{(d)}}(y_n^{(t)}) &= p_{\theta^{(d)}}(y_n^{(t)} | x_n^{(t)}), \label{eq:ibt1}
     \\
     p_{\theta^{(t)}}(x_m^{(d)}) &=  p_{\theta^{(t)}}(x_m^{(d)} | y_m^{(d)}).
    \label{eq:ibt2}
\end{align}
From a probabilistic perspective, iterative back-translation can be seen as successively updating the posterior of the missing data (resp. text) using the conditional likelihood of the observed data (resp. data) from the other model component. 

Note that under the iterative back-translation scheme, optimizing the log-likelihoods is equivalent to optimizing the cycle objective for unsupervised learning (see Appendix~\ref{sec:cycle_obj}) from prior work \citep{hoang-etal-2018-iterative, guo-etal-2020-cyclegt, DBLP:conf/aistats/GuoJ0Q0ZZW21}:
\begin{align}
    \mathcal{L}_{\mathrm{cycle}}(\theta^{(t)}, \theta^{(d)}) &= \underset{p_T(x)}{\mathbb{E}}\bigg[\underset{p_{\theta^{(d)}}(y | x)}{\mathbb{E}}[\log p_{\theta^{(t)}} (x | y)]\bigg] \notag\\
    &+ 
    \underset{p_D(y)}{\mathbb{E}}\bigg[\underset{p_{\theta^{(t)}}(x | y)}{\mathbb{E}}[\log p_{\theta^{(d)}} (y | x)]\bigg].
    \label{eq:cycle_obj}
\end{align}


\section{AUGMENTED LATENT VARIABLE MODEL FOR DATA-TO-TEXT AND TEXT-TO-DATA}
\label{sec:augmented-model}
Our model, shown in Fig.~\ref{fig:augpgm} and described below, is a refined version of the general scheme illustrated in Fig.~\ref{fig:nonaugpgm}. Because of the non-alignment assumption, the corresponding model leads to intractable expressions of the likelihood. Therefore, we use a
variational autoencoder (VAE) \citep{Kingma2014, DBLP:conf/icml/RezendeMW14} to make learning and inference tractable.
\subsection{Augmented Latent Variable Model}
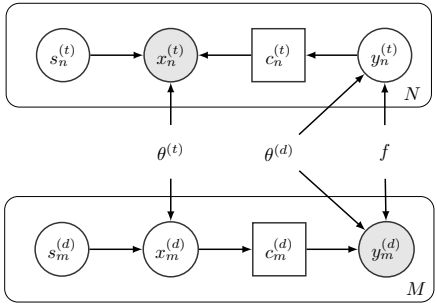
\begin{figure}
\centering
\resizebox{0.70\columnwidth}{!}{\begin{tikzpicture}
\newcommand{\mynodedistance}{5mm}
\newcommand{\mynodesize}{10mm}
\tikzstyle{observed}=[circle, minimum size = \mynodesize, thick, draw =black!80, fill = black!10, node distance = \mynodedistance]
\tikzstyle{latent}=[circle, minimum size = \mynodesize, thick, draw =black!80, node distance = \mynodedistance]
\tikzstyle{latent_invisible}=[circle, minimum size = \mynodesize,  node distance = \mynodedistance]
\tikzstyle{parameter}=[circle, minimum size = \mynodesize, draw =white, node distance = \mynodedistance]
\tikzstyle{non-random-variable}=[rectangle, minimum size = \mynodesize, draw =black!80, thick, node distance = \mynodedistance]
\tikzstyle{connect}=[-latex, thick]
\tikzstyle{plate}=[rectangle, draw=black!100, rounded corners=0.25cm]
\tikzstyle{plate_invisible}=[rectangle, rounded corners=0.25cm]
  \node[observed] (x_t) {$x_n^{(t)}$};
  \node[latent]   (s_t) [left=1cm of x_t] {$s_n^{(t)}$};
  \node[non-random-variable]   (c_t) [right=1cm of x_t] {$c_n^{(t)}$};
  \node[latent]   (y_t) [right=3cm of x_t]{$y_n^{(t)}$};
  \path (s_t) edge [connect] (x_t);
  \path (c_t) edge [connect] (x_t);
  \path (y_t) edge [connect] (c_t);
  \node[plate, 	 fit= (s_t)(y_t), inner sep=4.6mm, xshift=-1mm]() {}; 
  \node[rectangle, fit= (y_t), thick, inner sep=0mm, label=below right:$N$, xshift=-3mm, yshift=0.5mm] {}; 
  \node[parameter] (format_shared) [below=0.8cm of y_t]{$f$};
  \node[parameter] (text_model_shared) [below=0.8cm of x_t]{$\theta^{(t)}$};
  \path (format_shared) edge [connect] (y_t);
  \path (text_model_shared) edge [connect] (x_t);
  \node[latent] (x_d) [below=0.8cm of text_model_shared] {$x_m^{(d)}$};
  \node[latent]   (s_d) [left=1cm of x_d] {$s_m^{(d)}$};
  \node[non-random-variable] (c_d) [right=1cm of x_d] {$c_m^{(d)}$};
  \node[observed]   (y_d) [right=3cm of x_d]{$y_m^{(d)}$};
  \node[parameter] (data_model_shared) [above=0.8cm of c_d]{$\theta^{(d)}$};
  \path (s_d) edge [connect] (x_d);
  \path (x_d) edge [connect] (c_d);
  \path (text_model_shared) edge [connect] (x_d);
  \path (data_model_shared) edge [connect] (y_t);
  \path (data_model_shared) edge [connect] (y_d);
  \path (c_d) edge [connect] (y_d);
  \node[plate, 	 fit= (s_d)(y_d), inner sep=4.6mm, xshift=-1mm]() {}; 
  \node[rectangle, fit= (y_d), thick, inner sep=0mm, label=below right:$M$, xshift=-3mm, yshift=0.5mm] {}; 
  \path (format_shared) edge [connect] (y_d);
\end{tikzpicture}}
\caption{Graphical representation of our model. We augment the existing variables in Fig.~\ref{fig:nonaugpgm} with additional latent variables $s_n^{(t)}$ and $s_m^{(d)}$ to capture the diversity in the text, and condition the generation of the structured data on the data format. 
}
\label{fig:augpgm}
\end{figure}

\textbf{Probabilistic graphical model.}
In Section~\ref{sec:unsuplearning}, we assumed that the content from structured data can be expressed in multiple ways, leading to different textual surface realizations. This is modeled by conditioning the text generation on a latent variable denoted $s$ for \textit{style}. We make use of two such variables,  $s_n^{(t)}$ and $s_m^{(d)}$, respectively, for conditioning on observed text occurrences $x_n^{(t)}$ and latent texts $x_m^{(d)}$, as shown in Fig.~\ref{fig:augpgm}. These style variables will be useful for modeling the diversity of texts - thus potentially allowing us to generate diverse texts from a single piece of data at inference time, and for modeling the uncertainty of the association between text and data during training. Similarly, we introduce a parameter, denoted $f$, that specifies the targeted conversion format for a text. Since the target format is available at both test and training time, this is a fixed parameter and not a latent one.


 
\textbf{ELBO.} 
We choose to leverage the VAE framework for our model. We show that the cycle-training objective can be deduced from the graphical model in a way that is similar to an iteration of the Wake-Sleep algorithm (see Appendix \ref{sec:elbo_derivation}) \citep{cotterell+kreutzer.arxiv18}.
Therefore, we derive the evidence lower bound (ELBO) equations for the text $x$ and the structured data $y$ as:
\begin{align}
\mathcal{L}_{x}(\theta^{(t)}, \phi) &=
\underset{p_{\theta^{(d)}}(\hat{y} | x; f)}{\mathbb{E}}\Bigg[
    \underset{q_\phi(s | x)}{\mathbb{E}}\bigg[\log p_{\theta^{(t)}}(x | \mathrm{Enc}_\phi(\hat{y}), s)\bigg] 
    \Bigg] \notag\\
    &~~~-D_{KL}(q_\phi(s | x) \| p(s)),
    \label{eq:loss_vae_x}
\\
\mathcal{L}_{y}(\theta^{(d)}, \phi)
&=
\underset{p(s)}{\mathbb{E}}\Bigg[\underset{p_{\theta^{(t)}}(\hat{x} | y, s)}{\mathbb{E}}\bigg[
        \log p_{\theta^{(d)}}(y | \mathrm{Enc}_\phi(\hat{x}); f) 
        \bigg]
    \Bigg],
\label{eq:loss_vae_y}
\end{align}
where $\mathrm{Enc}_{\phi}$ is the sequence-to-sequence encoder used for learning representations of both text and data (see Section Content encoding). 

\textbf{Priors.}
We place a Gaussian prior over the style of the text,
$$
p(s_n^{(t)}) = p(s_m^{(d)}) = \mathcal{N}(0,I).
$$ 
Since we want to incorporate information from observed text and data into their corresponding latent variables, we follow the iterative back-translation scheme described in Section~\ref{sec:ibt}, Eqs.~\ref{eq:ibt1} and \ref{eq:ibt2}, and define the priors as
\begin{align}
p_{\theta^{(d)}}(x_m^{(d)}) &= p_{\theta^{(d)}}(x_m^{(d)} | y_m^{(d)}),\\
p_{\theta^{(t)}}(y_n^{(t)}) &= p_{\theta^{(t)}}(y_n^{(t)} | x_n^{(t)}).
\end{align}
\textbf{Variational posteriors.}
For the variational posterior of the style, we use different parameterizations for $s_n^{(t)}$ and $s_m^{(d)}$, since in the former case we have access to observed text $x_n^{(t)}$, while for the latter there is no style information in the observed $y_m^{(d)}$. We parameterize $q_{\phi}(s_n^{(t)} | x_n^{(t)})$ and $q_{\phi}(s_m^{(d)} | y_m^{(d)})$ with a family of Gaussian distributions:
    \begin{align}
        q_{\phi}(s_n^{(t)} | x_n^{(t)}) &= \mathcal{N}(\mu_{\phi}(x_n^{(t)}); \Sigma_{\phi}(x_n^{(t)})), \\
        q_{\phi}(s_m^{(d)} | y_m^{(d)}) &= \mathcal{N}(0,I).
    \end{align}

\textbf{Content encoding.}
\label{sec:content_enc}
We initially considered encoding the content stochastically using a family of Gaussian distributions as in \cite{Ye2020Variational}. However, we found in preliminary experiments that considering the content as stochastic leads to a performance decrease, especially when regularizing with a Gaussian prior. Therefore, we follow a simpler approach and choose to encode the content $c_n^{(t)}$ and $c_m^{(t)}$ deterministically by setting
\begin{align}
    c_n^{(t)} = \mathrm{Enc}_{\phi}(y_n^{(t)});~c_m^{(d)} = \mathrm{Enc}_{\phi}(x_m^{(d)}),
\end{align} where $\mathrm{Enc}_{\phi}$ is the encoder of the T5 model in our case.

\begin{figure}
    \centering
    \includegraphics[width=\columnwidth]{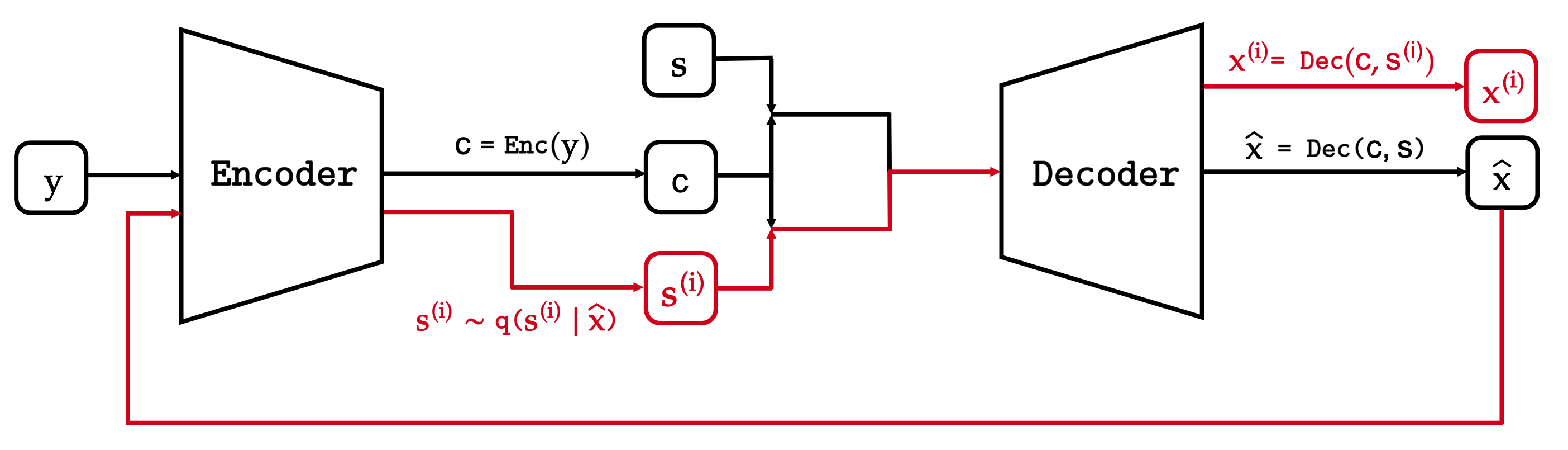}
    \caption{Given an observed data $y$, we generate a synthetic text $\hat{x}$ (black path). We use the synthetic text to retrieve the corresponding style posterior $q(s | \hat{x})$. We can now sample multiple style latent codes $s^{(i)} \sim q(s^{(i)} | \hat{x})$ and feed them into the decoder, along with the encoded content $c = \mathrm{Enc}(y)$, to generate texts with multiple styles $x^{(i)} = \mathrm{Dec}(c,s^{(i)})$ (red path).}
    \label{fig:d2t-inference}
    \vspace{-3pt}
\end{figure}

\textbf{Text-to-data inference.}
In order to infer data $y$ from text $x$, we first pass a one-hot encoded data format vector $f$ through a linear layer $s_f = W_f f + b_f$, and then feed it along with the encoded content $c = Enc_\phi(x)$ to the decoder to generate $y \sim p_\theta(y | c, s_f)$. 

\textbf{Data-to-text inference.}
Here we clarify how to infer text from data, i.e., $p_\theta(x | y)$. During training, the text style latent code is encoded from the text distribution, which we do not have at inference time. The usual way to infer text from data in this scenario is to sample from the prior, as the divergence term in the ELBO pushes the variational posterior to its respective prior. Empirically, we found that this sampling method is acceptable when the dimensionality of the latent code is low. However, in the case of a high-dimensional latent space, the model fails to learn. We suspect that this is due to the inherent discrete nature of text; the latent space of the VAE for text exhibits regions of low density and can contain multiple holes \citep{xu2020variational, lat_holes}, leading to the generation of incoherent text when naively decoding a latent code sampled from the prior. This effect is stronger in our case, since we need to perform inference starting at the very beginning of training. This results in noisy and uninformative samples, and leads to the failure of iterative back-translation.

In order to stabilize the inference process, when optimizing for structured data $y$, instead of sampling from the prior, we choose to fix the style latent code during training to $s = [0,...,0]$ (the mean of the Gaussian prior) in order to infer synthetic text via $\hat{x} \sim p_\theta(x | c, s)$, where $c = Enc_\phi(y)$.
At inference time, to mitigate the train-test discrepancy, we infer $p_\theta(x | y)$ by first generating $\hat{x} \sim p_\theta(x | c, s)$ with $s = [0, ... , 0]$ and $c = \mathrm{Enc}_\phi(y)$, and then use this synthetic sample to generate the style latent code $s \sim q_\phi(s | \hat{x})$. We can then infer $x \sim p_\theta(x | c, s)$ (see Fig.~\ref{fig:d2t-inference}).

\textbf{Disentanglement.} 
As we encode the style of the text and the content using the same encoder, it is likely that both variables are entangled. We introduce a way to control their degree of entanglement, since this may influence the performance of the model.
One way to assess the level of disentanglement between two random variables $X$ and $Y$ is by computing their Mutual Information (MI), and disentangling $X$ from $Y$ is achieved by minimizing their MI. While some works have explicitly sought to minimize this quantity or an estimate of it \citep{disen_sanchez, cheng2020club, cheng-etal-2020-improving, colombo-etal-2021-novel}, other works \citep{DBLP:conf/icml/WangSZRBSXJRS18, pmlr-v162-chang22a} have simply chosen to introduce a bottleneck on the capacity of the style encoder, constraining the dimensionality of one of the two variables in order to limit the information flow from one variable to the other, also known as the \textit{content leakage} phenomenon \citep{content_leakage}.
We follow the latter approach and show (see Appendix.~\ref{sec:disentanglement}) that we can upper-bound the MI between the content and the style variables with a bound that scales with the dimension of the style of the text and the structured data format variables. 

\subsection{Training}
\RestyleAlgo{ruled}
\definecolor{darkgreen}{RGB}{0,128,0}
\definecolor{darkorange}{RGB}{255,140,0}
\definecolor{darkblue}{RGB}{0,0,139}
\SetKwComment{Comment}{/* }{ */}
\SetKwInOut{Parameters}{Parameters}

\begin{algorithm}[hbt!]
\DontPrintSemicolon
\caption{Cycle-training via back-translation}\label{alg:bt_alg}
\KwIn{$\bs{X}^{(t)}$, $\bs{Y}^{(d)}$, format $f$, Decoder $\theta$, Encoder $\phi$}
\While{not converged}{
    $(x, y) \sim \left(\bs{X}^{(t)}, \bs{Y}^{(d)}\right)$\;
    \textcolor{darkgreen}{\tcp{Denoising objectives}}\relax
    $\nabla(\theta,\phi) \gets \nabla_{(\theta,\phi)} \mathcal{L}_{denoising}(\theta, \phi)$ \hspace*{\fill} (Eq.~\ref{eq:denoising_loss}) \;
    \textcolor{darkgreen}{\tcp{T2D ($y \rightarrow \hat{x} \rightarrow y $)}}\relax
    $s \sim \mathcal{N}(0,I)$ (or $s \gets [0,...,0]$)\;
    $\hat{x} \sim p_\theta(x \mid y,s)$ \textcolor{darkgreen}{\Comment*[r]{no gradient}}
    $c \gets Enc_\phi(\hat{x})$\;
    $\nabla(\theta,\phi) \gets \nabla(\theta,\phi) + \nabla_{(\theta,\phi)} \mathcal{L}_{y}(\theta, \phi)$ \hspace*{\fill} (Eq.~\ref{eq:loss_vae_y}) \;
    \textcolor{darkgreen}{\tcp{D2T ($x \rightarrow \hat{y} \rightarrow x$)}}\relax
    $\hat{y} \sim p_\theta(y \mid x,f)$ \textcolor{darkgreen}{\Comment*[r]{no gradient}}
    $c \gets Enc_\phi(\hat{y})$\;
    $\nabla(\theta,\phi) \gets \nabla(\theta,\phi) + \nabla_{(\theta,\phi)} \mathcal{L}_{x}(\theta, \phi)$ \hspace*{\fill} (Eq.~\ref{eq:loss_vae_x}) \;
    \textcolor{darkgreen}{\tcp{Update}}\relax
    $(\theta, \phi) \gets (\theta,\phi) + \nabla(\theta,\phi)$\; 
}
\end{algorithm}

One potential danger that can arise when only optimizing the cycle objective is a collapse onto the trivial identity mapping, which may happen when the source and target domains are the same. We encounter this phenomenon due to our linearization of the structured data, where we implicitly frame the D2T and T2D tasks as translation tasks. Learning the trivial identity mapping for both tasks is one way to locally maximize the objective.
To remedy this issue, \cite{lample2018unsupervised} and \cite{schmitt-etal-2020-unsupervised} introduce denoising objectives, which make the model robust to noise. We follow the same methodology as \cite{schmitt-etal-2020-unsupervised} and adapt their noise functions (\textit{swap}, \textit{drop}, \textit{blank}, \textit{repeat}, \textit{rule}) for all structured data formats (see Appendix~\ref{sec:noise_fn} for examples).

In addition to the ELBO, we also optimize the following denoising objective:
\begin{align}
    \mathcal{L}_{denoising}(\theta) &= \underset{x \sim p_T(x), \tilde{x} \sim Noise(x)}{\mathbb{E}}[\log p_\theta (x | \tilde{x})] \notag\\
    &+ \underset{y \sim p_D(y), \tilde{y} \sim Noise(y)}{\mathbb{E}}[\log p_\theta (y | \tilde{y})]. \label{eq:denoising_loss}
\end{align}

Iterative back-translation leverages the use of synthetic samples. When optimizing for a text $x$, we sample synthetic data $y \sim p_{\theta^{(d)}}(y | x)$, with $\theta^{(d)}$ being fixed, and we do the same with $\theta^{(t)}$ when optimizing for a structured data $y$. Since we are jointly optimizing Eq.~\ref{eq:loss_vae_x} and Eq.~\ref{eq:loss_vae_y} using the same model $\theta = (\theta_t, \theta_d)$, we accumulate gradients for text and data before updating the parameters of the model. Our method is described in Algorithm~\ref{alg:bt_alg}.

\section{EXPERIMENTS}
\begin{table*}[t!]
\caption{Evaluation results for D2T and T2D}
\label{tab:d2t_t2d_eval_results}
    \centering
    \begin{subtable}[h]{\textwidth}
    \centering
\label{tab:d2t_eval}
\tiny
\begin{tabular}{lrr|rr|rr|rr}\toprule
&\textbf{DART} & &\textbf{WEBNLG} & &\textbf{E2E} & &\textbf{TOTTO} & \\\toprule
&\textbf{BLEU} &\textbf{METEOR} &\textbf{BLEU} &\textbf{METEOR} &\textbf{BLEU} &\textbf{METEOR} &\textbf{BLEU} &\textbf{METEOR} \\\toprule
MS-Sup &\textbf{52.613} &\textbf{38.29} &\textbf{52.037} &\textbf{35.85} &40.679 &35.3 &\textbf{44.196} &\textbf{37.32} \\
SS-Sup &45.163 &35.91 &50.502 &35.58 &\textbf{40.95} &\textbf{35.43} &43.74 &37.08 \\\toprule
\rowcolor{lightgray} MS-UnSupVAE (ours) &\textbf{51.074} &\textbf{37.47} &\textbf{49.575} &\textbf{34.3} &39.33 &34.94 &\textbf{36.793} &\textbf{36.11} \\
SS-UnSupVAE &39.23 &33.35 &43.73 &32.5 &39.31 &35.02 &32.85 &33.83 \\
MS-UnSup &48.435 &36.2 &45.172 &31.72 &\textbf{39.529} &\textbf{34.95} &35.129 &34.92 \\
SS-Unsup &39.27 &33.19 &44.72 &32.59 &39.36 &34.79 &33.40 &34.21 \\
\bottomrule
\label{tab:D2Teval}
\end{tabular}
    \vspace{10pt}
    \end{subtable}
    \begin{subtable}[h]{\textwidth}
    \centering
\label{tab:t2d_eval}
\tiny
\begin{tabular}{lrrr|rrr|rrr|rrrr}\toprule
& &\textbf{DART} & & &\textbf{WEBNLG} & & &\textbf{E2E} & & &\textbf{TOTTO} & \\\toprule
&\textbf{ENT F1} &\textbf{REL F1} &\textbf{SemBLEU} &\textbf{ENT F1} &\textbf{REL F1} &\textbf{SemBLEU} &\textbf{ENT F1} &\textbf{REL F1} &\textbf{SemBLEU} &\textbf{ENT F1} &\textbf{REL F1} &\textbf{SemBLEU} \\\toprule
MS-Sup &\textbf{92.01} &\textbf{81.36} &\textbf{69.1} &\textbf{73.59} &\textbf{47.5} &\textbf{47.55} &\textbf{97.87} &\textbf{97.64} &\textbf{99.21} &\textbf{49.82} &\textbf{15.04} &11.9 \\
SS-Sup &84.78 &71.49 &54.5 &70.84 &45.46 &39.87 &96.97 &96.61 &98.85 &49.62 &14.03 &\textbf{12.82} \\\toprule
\rowcolor{lightgray} MS-UnSupVAE (ours) &\textbf{84.64} &\textbf{72.65} &\textbf{67.73} &\textbf{65.38} &\textbf{39.22} &\textbf{35.62} &\textbf{88.16} &\textbf{88} &\textbf{95.32} &\textbf{47.02} &\textbf{13.76} &\textbf{15.14} \\
SS-UnSupVAE &72.56 &56.03 &39.71 &57.22 &25.37 &16.09 &85.57 &85.18 &93.04 &43.36 &10.21 &5.72 \\
MS-UnSup &80.74 &67.69 &56.6 &60.45 &33.73 &26.97 &86.44 &86.13 &95.08 &44.95 &11.92 &10.19 \\
SS-Unsup &71.29 &55.12 &42.96 &57.3 &25.47 &13.62 &78.96 &78.59 &80.56 &44.26 &10.52 &8.06 \\
\bottomrule
\label{tab:T2Deval}
\end{tabular}

    \end{subtable}
    \vspace{-6pt}
\end{table*}

\subsection{Datasets}

We simulate a heterogeneous multi-source setting by considering a collection of D2T datasets. Each dataset uses a specific structured encoding. Since they are originally designed for a supervised setting, they are all aligned. We simply remove this alignment information and consider each source as a set of text and structured data. All the datasets are available from the GEM Benchmark \citep{gem_benchmark}. Due to the heterogeneous sizes of these datasets, we use a temperature mixing of $T=2$, as in \cite{T5_JMLR:v21:20-074}, to train our models in the multi-source setting.

\textbf{WebNLG 2020 (English)} \citep{castro-ferreira20:bilin-bi-direc-webnl-shared} is a dataset composed of 35,426 pairs of knowledge graphs and text crawled from 15 categories of DBpedia. This is follow-up work from the WebNLG 2017 challenge \citep{gardent2017creating}.

\textbf{Cleaned E2E} \citep{e2e_cleaned} results from cleaning the original E2E NLG dataset \citep{novikova-etal-2017-e2e}, an English benchmark dataset for data-to-text models that verbalizes a set of 2--9 key-value attribute pairs in the restaurant domain. The cleaned version has filtered out examples with hallucinations and outputs that do not fully cover all input attributes.

\textbf{DART} \citep{nan-etal-2021-dart} is an English dataset composed of 62,659 parallel text-data pairs, which aggregates multiple data-to-text datasets, including E2E and WebNLG, into a common format called data records. These data records consist of sets of triples and can be seen as a flat representation of the graph structure.

\textbf{ToTTo} \citep{parikh-etal-2020-totto} is a high-quality English table-to-text dataset with more than 120,000 examples, where each example is composed of a table from Wikipedia with highlighted cells that is paired with a sentence that describes these cells. 
All examples in this dataset were post-edited in multiple steps to ensure that the targets are fully faithful to the input information. 

\subsection{Baselines}
We consider the following baselines for our experiments (MS holds for multi-source, SS for single-source): 

\textbf{MS-Sup}: This pre-trained T5 model is fine-tuned on all datasets in a multi-source, \textit{supervised} setting. We train a \textit{separate model} for D2T and T2D. For T2D, we consider one prefix for each format and train in a multi-task fashion.

\textbf{SS-Sup}: This pre-trained T5 model is fine-tuned on each dataset independently and is \textit{supervised}. We train a separate model for D2T and T2D.

\textbf{MS-UnSupVAE}: Our approach.

\textbf{SS-UnSupVAE}: This model is a single pre-trained T5 model fine-tuned on each dataset separately for both D2T and T2D. This approach is close to the one in \cite{DBLP:conf/aistats/GuoJ0Q0ZZW21}, but they use separate models for D2T and T2D and additional preprocessing for T2D.

\textbf{MS-UnSup}: This pre-trained T5 model is fine-tuned on all datasets in a multi-source, \textit{unsupervised} setting. We consider a prefix for each format and train a single model for both D2T and T2D in a multitask fashion.

\textbf{SS-UnSup}: This model uses a pre-trained T5 model on each dataset separately for both D2T and T2D. This is comparable to \cite{schmitt-etal-2020-unsupervised}, which uses a BiLSTM-LSTM architecture as the encoder-decoder instead of a T5 model in our case.

Note that because of limited resources, we conducted all of our experiments using the \textit{T5-small} model. However, we will see that for some experiments we reach the performance of state-of-the-art \textit{T5-large} models.

\subsection{Evaluation metrics}
\textbf{D2T metrics.} We assess traditional metrics in our experiments, including BLEU \citep{papineni-etal-2002-bleu, sacrebleu}, ROUGE-L \citep{lin-2004-rouge}, and METEOR \citep{banerjee-lavie-2005-meteor, denkowski-lavie-2014-meteor}. 

\textbf{T2D metrics.} 
Little has been established in the literature for the evaluation of generated structured data. 
We follow the same approach as prior works on text-to-graph \citep{guo-etal-2020-cyclegt, DBLP:conf/aistats/GuoJ0Q0ZZW21, schmitt-etal-2020-unsupervised} and report the precision, recall, and f1 score of predicted entities and relations for all formats. In addition to these metrics, we also compute the SemBLEU score \citep{song-gildea-2019-sembleu} by converting all formats into a graph structure. This score, which is adapted to the structure of graphs, provides more information on higher-order n-grams than previous entity and relation scores.

\textbf{Diversity metrics.} Since our model has the ability to generate diverse texts from the same data, it is important to assess the effectiveness of this mechanism. We evaluate the diversity of the generated texts in our experiments using the Self-BLEU metric \citep{zhu2018texygen-self-bleu}, which treats one generated sentence as the hypothesis and the others as the reference, and calculates the BLEU score for every set of generated sentences. The score is then averaged over the test set. We also report Distinct-1 and Distinct-2 \citep{li-etal-2016-diversity-distinct}, which corresponds to the number of distinct unigrams and bigrams in generated texts.
This metric is then scaled by the total number of generated tokens, which penalizes long sentences.

\subsection{Results}
We present some of our results in Table~\ref{tab:d2t_t2d_eval_results}. A more complete list of results can be found in Table~\ref{tab:eval_results} (Appendix).

\textbf{D2T.} (i) Supervised vs Unsupervised. The supervised model \textbf{MS-Sup}, which is effectively an upper bound on model performance, performs the best as expected.
(ii) Single source vs multi-source. Except for E2E, multi-source models always outperform their single-source counterparts. Learning from multiple sources improves performance across datasets. Since the E2E dataset is a very specific, small, and closed domain (restaurant) dataset, training on multiple domains may introduce interference. (iii) VAE vs non VAE. For the multi-source setting, the VAE version \textbf{MS-UnSupVAE} always outperforms its non-VAE counterpart \textbf{MS-UnSup} by a significant margin. This result highlights the importance of uncertainty modeling when the conversion problem becomes more challenging, as is the case for multi-source vs single-source scenarios. (iv) Unsupervised models. Except on E2E, where all models have similar performance, our model outperforms all other unsupervised approaches across all metrics, and closes the performance gap with the supervised baselines. Note that \textbf{MS-UnSupVAE} is particularly effective on DART, outperforming even the single-source supervised baselines and current state-of-the-art obtained with larger models (see Appendix~\ref{tab:eval_results}).
We attribute this performance to the multi-source nature of DART. By flattening knowledge graphs into sets of triples, the data records in DART have lost information that was present in the original data. Overall, it appears that training on heterogeneous formats that originate from the same content improves model performance.

\textbf{T2D.} First, we notice that using a text-to-text model for T2D results in some format errors, although they are not frequent (less than 5\% for all datasets). Training on multiple sources is challenging for the T2D task, as it also increases format errors due to the consideration of several formats, such as for the \textbf{MS-Sup} model on the E2E dataset. Compared to their supervised counterparts, all unsupervised approaches underperform on entity and relation extraction. Nevertheless, our model remains competitive with the other supervised baselines.

\textbf{Diversity Results.}
\begin{figure}
    \centering
    \includegraphics[width=\linewidth]{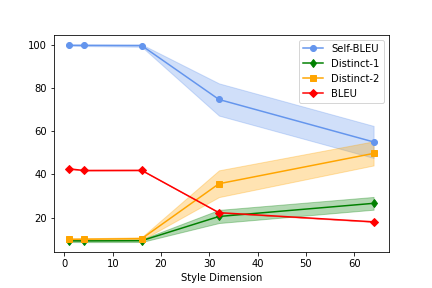}
    \caption{Evolution of average BLEU, Distinct-1, Distinct-2 (higher is better), and self-BLEU (lower is better) metrics over all datasets as the dimensionality of the style increases.}
    \label{fig:diversity_results}
\end{figure}
We jointly evaluate the diversity and quality of the text generated by our model as the dimension of the style latent variable varies. Due to the stochasticity of text generation, we report the mean and 95\% confidence intervals for the diversity metrics and the BLEU metrics (quality) for 10 different random seeds. We average the metrics over all datasets. We observe that increasing the dimensionality of the style vector increases the Distinct-(1,2) scores, i.e., the number of unigrams and bigrams increases. The generated sentences are also more diverse, since Self-BLEU decreases. It appears that there is little variation in the generation when the style dimensionality is fairly low (less than 16). 
As the dimensionality increases, however, the model tends to hallucinate, as the BLEU metrics generally decrease (see more examples in Appendix~\ref{sec:more_ex}). 

\section{ASSUMPTIONS AND LIMITATIONS}
One limitation of our work is that we have considered sources with structured data formats that can be mapped to sets of relational triples. The treatment of more diverse structured formats and tasks remains an open problem.
A second limitation concerns the diversity in the text domain and in the relation types considered in different corpora. Although our model behaves well on sources with different text domains and relations, the types of relations remain relatively simple, typical of the benchmarks in this domain, and extension to more complex situations is challenging. 
A potentially limiting assumption concerns our non-parallel framework for D2T and T2D. Although we remove alignment from the data by breaking aligned pairs, the domains of the two modalities remain aligned, which differs from the non-parallel framework considered in machine translation \citep{lample2018unsupervised}, where no alignment is assumed. This is due to the lack of large open-domain D2T datasets.

\section{CONCLUSION}
We have proposed a model capable of learning D2T and T2D mappings in a non-parallel setting, where text and data come from multiple sources. Our experiments show that the model is able to leverage this diversity to provide improved performance compared to unsupervised baselines. Our results suggest that mixing multiple formats can potentially improve model performance in both the parallel and non-parallel setting. The advantage of our method lies in its applicability to more realistic problems, where the data is heterogeneous, or where a strict match between text and data cannot be found. A promising avenue for future work would be to consider the adaptation to scenarios where new sources with scarce data are made available, requiring few-shot learning capability.

\subsubsection*{Acknowledgements}
The work has been partly funded through project ACDC ANR-21-CE23-0007.
We would like to express our sincere thanks to Alexandre Thomas who did an internship at Criteo and whose work served as the foundation for this paper.

\bibliographystyle{abbrvnat}
\bibliography{bibliography_shortnames} 


\clearpage
\appendix

\thispagestyle{empty}


\onecolumn

\section{APPENDIX}
\label{sec:appendix}
\subsection{Log-likelihood of the Latent Variable Model}
\label{sec:loglikelihood_latent_variable_model}
Given any arbitrary observed text $x_n^{(t)}$, we can express its log-likelihood using the missing data distribution $p_{\theta^{(d)}}(y_n^{(t)})$ as
\begin{align}
\log p_{\theta}(x_n^{(t)}) 
&= \underset{p_{\theta^{(d)}}(y_n^{(t)} )}{\mathbb{E}}[\log p_{\theta}(x_n^{(t)})]\notag\\
&= \underset{p_{\theta^{(d)}}(y_n^{(t)} )}{\mathbb{E}}\bigg[\log \frac{p_{\theta^{(t)}}(x_n^{(t)} | y_n^{(t)})p_{\theta^{(d)}}(y_n^{(t)} )}{p_{\theta^{(d)}}(y_n^{(t)}|x_n^{(t)})}\bigg] \notag\\
&= \underset{p_{\theta^{(d)}}(y_n^{(t)} )}{\mathbb{E}} [ \log p_{\theta^{(t)}}(x_n^{(t)} | y_n^{(t)})] + D_{KL}(p_{\theta^{(d)}}(y_n^{(t)} ) \| p_{\theta^{(d)}}(y_n^{(t)} | x_n^{(t)})).
 \label{eq:loglikelihood}
\end{align}

Likewise, given any observed data $y_m^{(d)}$, for any arbitrary density $p_{\theta^{(t)}}(x_m^{(d)})$ we have
\begin{align}
\log p_{\theta}(y_m^{(d)})&=\underset{p_{\theta^{(t)}}(x_m^{(d)} )}{\mathbb{E}} [ \log p_{\theta^{(d)}}(y_m^{(d)} | x_m^{(d)})] + D_{KL}(p_{\theta^{(t)}}(x_m^{(d)} ) \| p_{\theta^{(t)}}(x_m^{(d)} | y_m^{(d)})). \label{eq:loglikelihood2}
\end{align}

\subsection{Connection between the Cycle Objective and the Log-likelihood}
\label{sec:cycle_obj}
We use the following notation: $\bs{X}^{(t)}=\{x_1^{(t)},...,x_N^{(t)}\}$ and $\bs{Y}^{(d)} = \{y_1^{(d)},...,y_M^{(d)}\}$ are the sets of observed texts and structured data, respectively. $\bs{X}^{(d)}=\{x_1^{(d)},...,x_N^{(d)}\}$ and $\bs{Y}^{(t)} = \{y_1^{(t)},...,y_M^{(t)}\}$ are the sets of missing texts and structured data, respectively. We denote by $\theta = (\theta^{(t)}, \theta^{(d)})$ the set of parameters of the generative model. Considering the probablistic graphical model (PGM) in Fig.~\ref{fig:nonaugpgm}, we have
\begin{align}
    \log p_\theta(\bs{X}^{(t)}, \bs{Y}^{(d)}) &=\sum_N \log p_{\theta^{(t)}}(x_n^{(t)}) + \sum_M \log p_{\theta^{(d)}}(y_m^{(d)}) \notag \\
    &\approx N \underset{x_n^{(t)} \sim p_T(x)} {\mathbb{E}}[\log p_{\theta^{(t)}}(x_n^{(t)})] + M \underset{y_m^{(d)} \sim p_D(y)} {\mathbb{E}} [\log p_{\theta^{(d)}}(y_m^{(d)})] \label{eq:same_distrib} \\
    &= N\underset{x_n^{(t)} \sim p_T(x)} {\mathbb{E}}\bigg[\underset{p_{\theta^{(d)}}(y_n^{(t)} )}{\mathbb{E}} [ \log p_{\theta^{(t)}}(x_n^{(t)} | y_n^{(t)})] + D_{KL}(p_{\theta^{(d)}}(y_n^{(t)} ) \| p_{\theta^{(d)}}(y_n^{(t)} | x_n^{(t)}))\bigg] \notag \\
    &+ M\underset{y_m^{(d)} \sim p_T(y)} {\mathbb{E}}\bigg[\underset{p_{\theta^{(t)}}(x_m^{(d)} )}{\mathbb{E}} [ \log p_{\theta^{(t)}}(y_m^{(d)} | x_m^{(d)})] + D_{KL}(p_{\theta^{(t)}}(x_m^{(d)} ) \| p_{\theta^{(t)}}(x_m^{(d)} | y_m^{(d)}))]\bigg]
\end{align}

Iterative back-translation gives us the following: 
\begin{align}
     p_{\theta^{(d)}}(y_n^{(t)}) &= p_{\theta^{(d)}}(y_n^{(t)} | x_n^{(t)}), \\
     p_{\theta^{(t)}}(x_m^{(d)}) &=  p_{\theta^{(t)}}(x_m^{(d)} | y_m^{(d)}). 
\end{align} 
Under these assumptions, the KL terms $D_{KL}(p_{\theta^{(t)}}(x_m^{(d)} ) \| p_{\theta^{(t)}}(x_m^{(d)} | y_m^{(d)}))$ and $D_{KL}(p_{\theta^{(d)}}(y_n^{(t)} ) \| p_{\theta^{(d)}}(y_n^{(t)} | x_n^{(t)})) )$ vanish. 
In addition, if $N=M$, then
\begin{align}
    \operatorname*{argmax}_\theta \log p_\theta(\bs{X}^{(t)}, \bs{Y}^{(d)})
    &\approx \operatorname*{argmax}_{(\theta^{(t)}, \theta^{(d)})} \underset{x_n^{(t)} \sim p_T(x)} {\mathbb{E}}\bigg[\underset{p_{\theta^{(d)}}(y_n^{(t)} | x_n^{(t)})}{\mathbb{E}} [ \log p_{\theta^{(t)}}(x_n^{(t)} | y_n^{(t)})]\bigg] \notag \\
    &~\quad + \underset{y_m^{(d)} \sim p_T(y)} {\mathbb{E}}\bigg[\underset{p_{\theta^{(t)}}(x_m^{(d)} | y_m^{(d)})}{\mathbb{E}} [ \log p_{\theta^{(t)}}(y_m^{(d)} | x_m^{(d)})]\bigg]  \notag \\
    &= \operatorname*{argmax}_{(\theta^{(t)}, \theta^{(d)})} \mathcal{L}_{\mathrm{cycle}}(\theta^{(t)}, \theta^{(d)})
\end{align}

Thus, optimizing the cycle objective amounts to optimizing the log-likelihood.

\subsection{Derivation of the ELBO}
\label{sec:elbo_derivation}
We use the following notation: $\bs{X}^{(t)}=\{x_1^{(t)},...,x_N^{(t)}\}$ and $\bs{Y}^{(d)} = \{y_1^{(d)},...,y_M^{(d)}\}$ are the sets of observed texts and structured data, respectively. $\bs{X}^{(d)}=\{x_1^{(d)},...,x_N^{(d)}\}$ and $\bs{Y}^{(t)} = \{y_1^{(t)},...,y_M^{(t)}\}$ are the sets of missing texts and structured data, respectively. $\bs{C}^{(t)} = \{c_1^{(t)},...,c_N^{(t)} \}$ and $\bs{C}^{(d)} = \{c_1^{(d)},...,c_M^{(d)}\}$ are the set of variables that encode the content from the missing text and structured data dataset, respectively. $\bs{S}^{(t)} = \{s_1^{(t)},...,s_N^{(t)} \}$ and $\bs{S}^{(d)} = \{s_1^{(d)},...,s_M^{(d)}\}$ are the set of latent variables that encode the styles of the text in $\bs{X}^{(t)}$ and $\bs{X}^{(d)}$, respectively. $f$ is the structured data format common to all $\bs{Y}^{(d)}$. We denote by $\theta = (\theta^{(t)}, \theta^{(d)})$ the set of parameters of the generative model and by $\phi$ the parameters of the variational posterior $q_\phi$. 

The joint likelihood of the model can be expressed as 
\begin{align}
    p_\theta(\bs{X}^{(t)}, \bs{Y}^{(d)}, \bs{X}^{(d)}, \bs{Y}^{(t)}, \bs{S}^{(t)}, \bs{S}^{(d)} ; f) \notag
    =~&\prod_{n=1}^N
    p(s_n^{(t)})p_{\theta^{(t)}}(x_n^{(t)} | s_n^{(t)}, c_n^{(t)})p_{\theta^{(d)}}(y_n^{(t)}; f)\notag\\
    &\prod_{m=1}^M
    p(s_m^{(d)})p_{\theta^{(t)}}(x_m^{(d)} | s_m^{(d)})p_{\theta^{(d)}}(y_m^{(d)} | c_m^{(d)} ; f).
\end{align}
Our goal is to maximize the marginal likelihood (evidence) $p_\theta(\bs{X}^{(t)}, \bs{Y}^{(d)};f)$ of the observed datasets. We can derive the Evidence Lower Bound (ELBO) as
\begin{align}
\log p_\theta(\bs{X}^{(t)}, \bs{Y}^{(d)} ; f)
&= 
\log \int p_\theta(\bs{X}^{(t)}, \bs{Y}^{(d)}, \bs{X}^{(d)}, \bs{Y}^{(t)}, \bs{S}^{(t)}, \bs{S}^{(d)} ; f)
\text{d}\bs{X}^{(d)}\text{d}\bs{Y}^{(t)}\text{d}\bs{S}^{(t)}\text{d}\bs{S}^{(d)} \notag \\
&=
\log \mathbb{E}_{q_\phi(\bs{X}^{(d)}, \bs{Y}^{(t)}, \bs{S}^{(t)}, \bs{S}^{(d)} | \bs{X}^{(t)}, \bs{Y}^{(d)} ; f)}
\left[
\frac{p_\theta(\bs{X}^{(t)}, \bs{Y}^{(d)}, \bs{X}^{(d)}, \bs{Y}^{(t)}, \bs{S}^{(t)}, \bs{S}^{(d)} ; f)}
{q_\phi(\bs{X}^{(d)}, \bs{Y}^{(t)}, \bs{S}^{(t)}, \bs{S}^{(d)} | \bs{X}^{(t)}, \bs{Y}^{(d)} ; f)}
\right]\notag\\
&\geq
\mathbb{E}_{q_\phi(\bs{X}^{(d)}, \bs{Y}^{(t)}, \bs{S}^{(t)}, \bs{S}^{(d)} | \bs{X}^{(t)}, \bs{Y}^{(d)};f)}
\left[
\log p_\theta(\bs{X}^{(t)}, \bs{Y}^{(d)} | \bs{X}^{(d)}, \bs{Y}^{(t)}, \bs{S}^{(t)}, \bs{S}^{(d)}; f)
\right]
\notag\\
&
\quad -
D_{KL}
\left(
q_\phi(\bs{X}^{(d)}, \bs{Y}^{(t)}, \bs{S}^{(t)}, \bs{S}^{(d)} | \bs{X}^{(t)}, \bs{Y}^{(d)};f) \| 
p_\theta(\bs{X}^{(d)}, \bs{Y}^{(t)}, \bs{S}^{(t)}, \bs{S}^{(d)}; f
\right) \notag\\
&~ \quad \text {via Jensen's inequality } \label{eq:elbo_jensen} \\
&= 
\mathcal{L}_{(\bs{X}^{(t)}, \bs{Y}^{(d)})}(\theta, \phi).
\end{align}
Let us now factorize the different terms. Given the PGM in Fig.~\ref{fig:augpgm}, the likelihood can be factorized as
\begin{align}
    p_\theta(\bs{X}^{(t)}, \bs{Y}^{(d)} | \bs{X}^{(d)}, \bs{Y}^{(t)}, \bs{S}^{(t)}, \bs{S}^{(d)} ; f) 
&= \prod_N p_{\theta^{(t)}}(x_n^{(t)} | s_n^{(t)}, \mathrm{Enc}_\phi(y_n^{(t)}))\prod_M p_{\theta^{(d)}}(y_m^{(d)} | \mathrm{Enc}_\phi(x_m^{(d)}) ; f).
\label{eq:likelihood_factorization}
\end{align}
The prior can be factorized as
\begin{align}
    p_\theta(\bs{X}^{(d)}, \bs{Y}^{(t)}, \bs{S}^{(t)}, \bs{S}^{(d)}; f) 
&= \prod_N p(s_n^{(t)})p_{\theta^{(d)}}(y_n^{(t)} ; f) \prod_M p(s_m^{(d)}) p_{\theta^{(t)}}(x_m^{(d)} | s_m^{(d)}).
\label{eq:prior_factorization}
\end{align}
And the variational posterior as
\begin{align}
    q_\phi(\bs{X}^{(d)}, \bs{Y}^{(t)}, \bs{S}^{(t)}, \bs{S}^{(d)} | \bs{X}^{(t)}, \bs{Y}^{(d)} ; f) 
    &= 
    \prod_N 
      q_\phi(s_n^{(t)} | x_n^{(t)})
      q_\phi(y_n^{(t)} | x_n^{(t)};f) 
    \prod_M 
      q_\phi (s_m^{(d)} | y_m^{(d)}) 
      q_\phi(x_m^{(d)} | s_m^{(d)};y_m^{(d)}).
    \label{eq:stable_posterior_factorization}
\end{align}
Note that Eq.~\ref{eq:elbo_jensen} holds for any arbitrary $q_\phi$; different factorizations of the variational posteriors are possible, for instance $q_\phi(\bs{X}^{(d)}, \bs{S}^{(d)} | \bs{X}^{(t)}, \bs{Y}^{(d)}) = q_\phi(\bs{X}^{(d)} | \bs{Y}^{(d)})q_\phi(\bs{S}^{(d)} | \bs{X}^{(d)})$, which would result in encoding the style with the synthetic text. We ran experiments with this variant and found that the model performs the best with Eq.~\ref{eq:stable_posterior_factorization}.

Following Eqs.~\ref{eq:likelihood_factorization},~\ref{eq:prior_factorization}, and~\ref{eq:stable_posterior_factorization}, The ELBO can be factorized as
\begin{align}
    \mathcal{L}_{(\bs{X}^{(t)}, \bs{Y}^{(d)})}(\theta, \phi) = \sum_N \mathcal{L}_{x_n^{(t)}}(\theta^{(t)}, \phi) + \sum_M \mathcal{L}_{y_m^{(d)}}(\theta^{(d)}, \phi),
\end{align}
where each term can be written as
\begin{align}
    \mathcal{L}_{x_n^{(t)}}(\theta^{(t)}, \phi) &= \underset{q_\phi(y_n^{(t)} | x_n^{(t)};f)}{\mathbb{E}}\bigg[\underset{q_\phi(s_n^{(t)} | x_n^{(t)})}{\mathbb{E}}[\log p_{\theta^{(t)}}(x_n^{(t)} | s_n^{(t)}; \mathrm{Enc}_\phi(y_n^{(t)}))]\bigg] \notag \\
    &-D_{KL}(q_\phi(y_n^{(t)} | x_n^{(t)} ; f) \| p_{\theta^{(d)}}(y_n^{(t)} ; f)) -D_{KL}(q_\phi(s_n^{(t)} | x_n^{(t)}) \| p(s_n^{(t)})), \\
    \mathcal{L}_{y_m^{(d)}}(\theta^{(d)}, \phi) &=\underset{q_\phi(s_m^{(d) } | y_m^{(d)})}{\mathbb{E}}\bigg[\underset{q_\phi(x_m^{(d)} | s_m^{(d)};y_m^{(d)})}{\mathbb{E}}[\log p_{\theta^{(d)}}(y_m^{(d)} | \mathrm{Enc}_\phi(x_m^{(d)}) ; f)] -D_{KL}(q_\phi(x_m^{(d)}|y_m^{(d)}) \| p_{\theta^{(t)}}(x_m^{(d)}| s_m^{(d)}))\bigg] \notag\\ &-D_{KL}(q_\phi(s_m^{(d)} | y_m^{(d)}) \| p(s_m^{(d)})).
\end{align}
Given our choices of parametrization and prior, we have $q_\phi(s_m^{(d)} | y_m^{(d)}) = p(s_m^{(d)}) = \mathcal{N}(0,I)$. As a consequence, the KL term $D_{KL}(q_\phi(s_m^{(d)} | y_m^{(d)}) \| p(s_m^{(d)}))$ vanishes.

We follow the iterative back-translation scheme in \cite{hoang-etal-2018-iterative} and consider the prior for the missing text (resp. data) distribution to be the conditional likelihood of the observed text (resp. data), i.e., 
\begin{align}
    p_{\theta^{(d)}}(y_n^{(t)} ; f) &= p_{\theta^{(d)}}(y_n^{(t)} | \mathrm{Enc}_\phi(x_n^{(t)}) ; f),\notag\\
    p_{\theta^{(t)}}(x_m^{(d)}| s_m^{(d)}) &= p_{\theta^{(t)}}(x_m^{(d)} | s_m^{(d)}, \mathrm{Enc}_\phi(y_m^{(d)})).
\end{align}

However, there is still one major obstacle that prevents us from directly optimizing these lower bounds. Indeed, the operations $y_n^{(t)} \sim q_\phi(y_n^{(t)} | f;x_n^{(t)})$ and $x_m^{(d)} \sim q_\phi(x_m^{(d)} | s_m^{(d)};y_m^{(d)})$ which correspond to the generation of synthetic training data for back-translation, is a sequence of discrete latent variables, and as a consequence cannot be easily reparametrized. To derive a tractable objective, we will assume that while we optimize for the text likelihood, the variational posterior is a good approximation of the conditional likelihood on the observed data, and vice-versa:
\begin{align}
    q_\phi(y_n^{(t)} | x_n^{(t)} ; f) &\approx p_{\theta^{(d)}}(y_n^{(t)} | \mathrm{Enc}_\phi(x_n^{(t)}) ; f), \label{eq:sleep1} \\
    q_\phi(x_m^{(d)} | s_m^{(d)}, y_m^{(d)}) &\approx p_{\theta^{(t)}}(x_m^{(d)} | s_m^{(d)},\mathrm{Enc}_\phi(y_m^{(d)})).
    \label{eq:sleep2}
\end{align}
This resonates with \cite{cotterell+kreutzer.arxiv18}, who interpreted iterative back-translation as iterations of the Wake-Sleep algorithm, where one alternatively learns a forward model $p_\theta(y | x)$ and a variational approximation $q_\phi(y | x)$. The only difference here is that we only consider the wake phase and choose to skip the sleep phase by assuming that Eqs.~\ref{eq:sleep1} and \ref{eq:sleep2} hold. This
assumption may be rather strong, but can be justified in the iterative back-translation framework, since both models $p_{\theta^{(t)}}(x | y)$ and $p_{\theta^{(d)}}(y | x)$ are
optimized to approximate the true conditional distributions $p_T(x | y)$ and $p_D((y | x)$, in alternate steps. Under these assumptions, the KL terms $D_{KL}(q_\phi(y_n^{(t)} | x_n^{(t)} ; f) \| p_{\theta^{(d)}}(y_n^{(t)} ; f))$ and $D_{KL}(q_\phi(x_m^{(d)}|y_m^{(d)}) \| p_{\theta^{(t)}}(x_m^{(d)}))$ vanish.
This results in these tractable bounds:
\begin{align}
    \mathcal{L}_{x_n^{(t)}}(\theta^{(t)}, \phi) &\approx \underset{p_{\theta^{(d)}}(y_n^{(t)} | \mathrm{Enc}_\phi(x_n^{(t)}) ; f)}{\mathbb{E}}\bigg[\underset{q_\phi(s_n^{(t)} | x_n^{(t)})}{\mathbb{E}}[\log p_{\theta^{(t)}}(x_n^{(t)} | s_n^{(t)}; \mathrm{Enc}_\phi(y_n^{(t)}))]\bigg] -D_{KL}(q_\phi(s_n^{(t)} | x_n^{(t)}) \| p(s_n^{(t)})), \\
    \mathcal{L}_{y_m^{(d)}}(\theta^{(d)}, \phi) &\approx \underset{p(s_m^{(d)})}{\mathbb{E}}\bigg[\underset{p_{\theta^{(t)}}(x_m^{(d)} | s_m^{(d)},\mathrm{Enc}_\phi(y_m^{(d)}))}{\mathbb{E}}[\log p_{\theta^{(d)}}(y_m^{(d)} | \mathrm{Enc}_\phi(x_m^{(d)}) ; f)] \bigg].
\end{align}
Note that synthetic samples $y_n^{(t)} \sim p_{\theta^{(d)}}(y_n^{(t)} | \mathrm{Enc}_\phi(x_n^{(t)}) ; f)$ and $x_m^{(d)} \sim p_{\theta^{(t)}}(x_m^{(d)} | s_m^{(d)},\mathrm{Enc}_\phi(y_m^{(d)}))$ used for back-translation are computed without gradients. When optimizing for $\theta^{(t)}$, the parameters $\theta^{(d)}$ resulting from the computation of $x_m^{(d)}$ are not updated, and vice-versa.

\subsection{Disentanglement through constraining the amount of information in style variables}
\label{sec:disentanglement}
Some works have defined a measure of disentanglement between two random variables $X$ and $Y$ as the Mutual Information (MI) between the two:
$$
I(X;Y) =  \mathbb{E}_{p(X,Y)} \left[\log \frac{p(X,Y)}{p(X)p(Y)} \right] = H(X) - H(X | Y) = H(Y) - H(Y | X)
$$
From a MI perspective, disentangling $X$ from $Y$ is achieved by minimizing their MI. While some works have explicitly sought to minimize this quantity or an estimate of it \citep{disen_sanchez, cheng2020club, cheng-etal-2020-improving, colombo-etal-2021-novel}, other works \citep{DBLP:conf/icml/WangSZRBSXJRS18, pmlr-v162-chang22a} have simply chosen to introduce a bottleneck on the capacity of the style encoder, constraining the dimensionality of one of the two variables in order to limit the information flow from one variable to the other. For completeness, we show here that by constraining the dimensionality of $s$, we can upper-bound the mutual information between the content $c$ and the style variables with a function that scales with the dimensionality.
\begin{itemize}
    \item For the style of the text, from the definition of MI, as the entropy is always positive, we can upper-bound it by:
    \begin{align}
        I(s; c) &\leq H(s) \notag\\
        &= H(\mathcal{N}(0,\boldsymbol{I})) \notag\\
        &= \frac{d}{2}(1+\log (2 \pi))+\frac{1}{2} \log |\boldsymbol{I}| = \frac{d}{2}(1+\log (2 \pi)),
    \end{align}
    where $d$ denotes the dimension of the multivariate Gaussian. Hence, by limiting the dimension of $s$, we prevent information flowing from the content variable $c$ to $s$.
    \item For the format $f$, as it is a parameter, the entropy is $0$:
    \begin{align}
        I(f; c) &\leq H(f) = 0.
    \end{align}
\end{itemize}
Note that although these bounds are not tight, this motivates the use of a lower dimensional space to encode the style.


\subsection{Examples from the "Rule" Noise Functions }
\label{sec:noise_fn}
\paragraph{T2D}
\begin{spverbatim}
Roadside Attractions distributed Super Capers which was also distributed by 
the public company Lionsgate.

KG: [HEAD] Attractions [TYPE] distributed [TAIL] Super [HEAD] also [TYPE] 
distributed [TAIL] public [HEAD] company [TYPE] has_attribute [TAIL] public

TRIPLESET: Attractions : distributed : Super | 
also : distributed : public | company : has_attribute : public

MR: name[Attractions], distributed[Super], name[also], 
distributed[public], name[company], has_attribute[public]

Totto: <page_title> </page_title> <section_title> </section_title> 
<table> <col_header> Entities </col_header> <col_header> Attractions </col_header> 
<col_header> also </col_header> <col_header> company </col_header> 
<cell> Super <col_header> distributed </col_header> 
<row_header> Attractions </row_header> </cell> 
<cell> public <col_header> distributed </col_header> 
<row_header> also </row_header> </cell> 
<cell> public <col_header> has_attribute </col_header> 
<row_header> company </row_header> </cell> </table>
\end{spverbatim}

\paragraph{D2T}
\begin{spverbatim}
[HEAD] Nord (Year of No Light album) [TYPE] artist [TAIL] Year of No Light 
[HEAD] Nord (Year of No Light album) [TYPE] genre [TAIL] Post-metal 
[HEAD] Nord (Year of No Light album) [TYPE] record label [TAIL] Crucial Blast
--> Nord (Year of No Light album) artist Year of No Light and 
Nord (Year of No Light album) record label Crucial Blast and 
Nord (Year of No Light album) genre Post-metal

ARA Veinticinco de Mayo (V-2) : length : 192000.0 (millimetres) | 
ARA Veinticinco de Mayo (V-2) : country : Argentina
--> ARA Veinticinco de Mayo (V-2) length 192000.0 (millimetres) and 
ARA Veinticinco de Mayo (V-2) country Argentina

name[The Phoenix], eatType[pub], food[French], priceRange[more than £30], 
customer rating[5 out of 5], area[riverside], familyFriendly[no], 
near[Crowne Plaza Hotel]
--> The Phoenix eatType pub and The Phoenix priceRange more than £30 and 
The Phoenix area riverside and The Phoenix near Crowne Plaza Hotel and 
The Phoenix name The Phoenix and The Phoenix customer rating 5 out of 5 and 
The Phoenix food French and The Phoenix familyFriendly no

<page_title> Gennady Golovkin vs. Daniel Jacobs </page_title> 
<section_title> CompuBox stats </section_title> 
<table> <cell> Golovkin <col_header> Fighter </col_header> </cell> 
<cell> 231/615 <col_header> Total punches </col_header> </cell> 
<cell> Jacobs <col_header> Fighter </col_header> </cell> 
<cell> 175/541 <col_header> Total punches </col_header> </cell> </table>
--> Gennady Golovkin vs. Daniel Jacobs and CompuBox stats and Golovkin Fighter and 
231/615 Total punches and Jacobs Fighter and 175/541 Total punches
\end{spverbatim}

\subsection{Implementation details}

\textbf{Adapting T5.}
We build our model around T5 \citep{T5_JMLR:v21:20-074}, which has an architecture that is close to the original Transformer architecture \citep{vaswani2017attention}.

We fine-tune the pre-trained \textit{t5-small} model to generate both text and data. We specify the target format by either adding the prefix "\textit{Generate text:}" (for $\theta^{(t)}$) or "\textit{Generate data:}" (for $\theta^{(d)}$) to the input sequence given to the encoder.

\textbf{Encoder-Decoder architectures.}
For the text style encoder $q_{\phi}(s | x)$, we add a special token "[STYLE]" to the text sequence that is passed to the encoder. We take the latent representation from the encoder output $h(x)$, and add a linear layer to obtain the parameters of our variational distribution:
$$\left\{
\begin{aligned}
    \mu(x) = W_\mu h(x) + b_\mu \\
    \Sigma(x) = W_\Sigma h(x) + b_\Sigma 
\end{aligned}
\right.$$
where $q_{\phi}(s | x) = \mathcal{N}(s; \mu(x), \Sigma(x))$.

For the structured data format $f$, we simply encode it using a linear layer via $s_f = W_f f + b_f $, before feeding it along with the encoded content $c = \mathrm{Enc}_\phi(x)$ to the decoder.
    
\textbf{KL Vanishing Issue.} This issue, also known as KL collapse, is a typical problem encountered when training VAE models. Fitting the variational posterior to the prior $\forall x, q_\phi(z | x) = p_\theta(z))$ to get a zero KL divergence term at the beginning of training is a simple way to minimize the ELBO loss. This results in a local minimum where the decoder is capable of performing effectively without the latent variable, which is typically the case with large language models and auto-regressive decoding \citep{fu-etal-2019-cyclical}. Many works have suggested fixes for this problem; we found the MMD-VAE approach \citep{DBLP:journals/corr/ZhaoSE17b} particularly effective.

In practice, the MMD-VAE approach replaces the KL divergence term $D_{KL}(q_\phi(z | x) \| p_\theta(z))$ with a sample estimate of the Maximum-Mean Discrepancy (MMD) measure between the aggregated posterior $q_\phi(z) = \underset{x\sim p_T(x)}{\mathbb{E}}[q_\phi(z | x)]$ and the prior $p_\theta(z)$.
We also experimented with the $\beta$-VAE and the cyclical schedule proposed in \cite{fu-etal-2019-cyclical} and used in \cite{li-etal-2020-optimus}, but preliminary experiments were not as successful.

\textbf{Hyperparameters} All models have been trained with the AdamW \citep{loshchilov2018decoupled} optimizer with a learning rate of $10^{-4}$ for $10$ epochs. Following \cite{lample2018unsupervised} and \cite{schmitt-etal-2020-unsupervised}, we set $p_{\mathrm{blank}} = p_{\mathrm{repeat}} = 0.2$, $p_{\mathrm{drop}} = 0.1$, and use all 5 noise functions \textit{swap}, \textit{drop}, \textit{blank}, \textit{repeat}, and \textit{rule}. For inference, we use beam search decoding for D2T and T2D with $5$ and $8$ beams, respectively. We use the coefficient $\lambda = 10$ for the MMD regularization term and use the traditional Gaussian kernel.

\begin{table*}\centering
\caption{D2T \& T2D Evaluation}\label{tab:eval_results}
\scriptsize
\begin{tabular}{lccc|ccc|ccc|cc}\toprule
&\textbf{} &\textbf{D2T} & & &\textbf{} & & &\textbf{T2D} & & & \\
&\textbf{} & & &\textbf{} &\textbf{ENTITY} & & &\textbf{RELATION} & & & \\
&\textbf{BLEU} &\textbf{METEOR} &\textbf{ROUGE-L} &\textbf{Precision} &\textbf{Recall} &\textbf{F1} &\textbf{Precision} &\textbf{Recall} &\textbf{F1} &\textbf{SemBLEU} &\textbf{Error (\%)} \\
\textbf{DART} & & & & & & & & & & & \\
T5-Large & 50.66 & \textbf{40.00} & - & - & - & - & - & - & - & - & - \\
BART-large & 48.56 & 39.00 & - & - & - & - & - & - & - & - & - \\
MS-Sup &\textbf{52.613} & 38.29 &\textbf{45.48} &\textbf{92.87} &\textbf{91.16} &\textbf{92.01} &\textbf{82.33} &\textbf{80.41} &\textbf{81.36} &\textbf{69.1} &0.2747 \\
SS-Sup &45.163 &35.91 &43.23 &85.66 &83.91 &84.78 &72.46 &70.54 &71.49 &54.5 &\textbf{0.2354} \\\toprule
\rowcolor{lightgray} MS-UnSupVAE &\textbf{51.074} &\textbf{37.47} &\textbf{44.77} &\textbf{85.17} &\textbf{84.11} &\textbf{84.64} &\textbf{73.18} &\textbf{72.12} &\textbf{72.65} &\textbf{67.73} &\textbf{0.2354} \\
SS-UnSupVAE &39.23 &33.35 &40.72 &73.89 &71.29 &72.56 &57.24 &54.86 &56.03 &39.71 &1.02 \\
MS-UnSup &48.435 &36.2 &43.71 &82.98 &78.63 &80.74 &70.12 &65.43 &67.69 &56.6 &0.4905 \\
SS-Unsup &39.27 &33.19 &40.51 &73.79 &68.96 &71.29 &57.65 &52.8 &55.12 &42.96 &0.706 \\ \\
\textbf{WEBNLG} & & & &\textbf{} & & & & & & & \\
MS-Sup &\textbf{52.037} &\textbf{35.85} &\textbf{39.53} &\textbf{73.71} &\textbf{73.48} &\textbf{73.59} &\textbf{48.16} &\textbf{46.85} &\textbf{47.5} &\textbf{47.55} &\textbf{1.18} \\
SS-Sup &50.502 &35.58 &39.31 &71.12 &70.57 &70.84 &46.24 &44.71 &45.46 &39.87 &0.5 \\\toprule
\rowcolor{lightgray} MS-UnSupVAE &\textbf{49.575} &\textbf{34.3} &\textbf{38.46} &\textbf{65.94} &\textbf{64.84} &\textbf{65.38} &\textbf{39.9} &\textbf{38.57} &\textbf{39.22} &\textbf{35.62} &4.834 \\
SS-UnSupVAE &43.73 &32.5 &37.14 &60.19 &54.53 &57.22 &27.59 &23.48 &25.37 &16.09 &5.06 \\
MS-UnSup &45.172 &31.72 &36.81 &62.35 &58.66 &60.45 &35.52 &32.12 &33.73 &26.97 &\textbf{1.237} \\
SS-Unsup &44.72 &32.59 &36.25 &60.5 &54.42 &57.3 &27.94 &23.39 &25.47 &13.62 &3.766 \\ \\
\textbf{E2E} & & & &\textbf{} & & & & & & & \\
MS-Sup &40.679 &35.3 &41.39 &\textbf{97.7} &\textbf{98.04} &\textbf{97.87} &\textbf{97.47} &\textbf{97.81} &\textbf{97.64} &\textbf{99.21} &0.1083 \\
SS-Sup &\textbf{40.95} &\textbf{35.43} &\textbf{41.48} &96.64 &97.29 &96.97 &96.27 &96.95 &96.61 &98.85 &\textbf{0.000} \\\toprule
\rowcolor{lightgray} MS-UnSupVAE &39.33 &34.94 &41 &\textbf{86.84} &\textbf{89.51} &\textbf{88.16} &\textbf{86.68} &\textbf{89.36} &\textbf{88} &\textbf{95.32} &\textbf{0.000} \\
SS-UnSupVAE &39.31 &35.02 &41.05 &86.56 &84.6 &85.57 &86.15 &84.22 &85.18 &93.04 &\textbf{0.000} \\
MS-UnSup &\textbf{39.529} &\textbf{34.95} &\textbf{41.28} &86.2 &86.67 &86.44 &85.88 &86.37 &86.13 &95.08 &\textbf{0.000} \\
SS-Unsup &39.36 &34.79 &41.01 &85.38 &73.44 &78.96 &84.97 &73.10 &78.59 &80.56 &\textbf{0.000} \\ \\
\textbf{TOTTO} & & & &\textbf{} & & & & & & & \\
MS-Sup &\textbf{44.196} &\textbf{37.32} &\textbf{36.65} &57.53 &\textbf{43.94} &\textbf{49.82} &\textbf{18.03} &\textbf{12.9} &\textbf{15.04} &11.9 &1.312 \\
SS-Sup &43.74 &37.08 &36.48 &\textbf{59.46} &42.57 &49.62 &17.76 &11.59 &14.03 &12.82 &\textbf{0.2727} \\\toprule
\rowcolor{lightgray} MS-UnSupVAE &\textbf{36.793} &\textbf{36.11} &\textbf{35.92} &54.57 &\textbf{41.3} &\textbf{47.02} &\textbf{16.56} &\textbf{11.77} &\textbf{13.76} &\textbf{15.14} &1.325 \\
SS-UnSupVAE &32.85 &33.83 &34.37 &\textbf{55.28} &35.67 &43.36 &14.02 &8.028 &10.21 &5.72 &\textbf{0.5065} \\
MS-UnSup &35.129 &34.92 &35.09 &54.34 &38.33 &44.95 &15.13 &9.827 &11.92 &10.19 &0.8961 \\
SS-Unsup &33.40 &34.21 &34.61 &55.41 &36.84 &44.26 &14.10 &8.38 &10.52 &8.06 &0.5714 \\
\bottomrule
\end{tabular}
\end{table*}

\subsection{More examples of generated D2T}
\label{sec:more_ex}
\textbf{Input:}
\begin{spverbatim}
[TABLECONTEXT] : title : American Playhouse | American Playhouse : role : Jed Jenkins | [TABLECONTEXT] : [title] : Jeff Daniels | American Playhouse : year : 1982
\end{spverbatim}

\textbf{Expected Ouput:}
\begin{spverbatim}
In 1982, Jed Jenkins in American Playhouse was played by Jeff Daniels.
\end{spverbatim}

\textbf{Generated Texts}:
\begin{spverbatim}
Jed Jenkins was a American Playhouse, serving from 1982 to 1982.
The American Playhouse is a play from Jed Jenkins to be cast in 1982.
Jed Jenkins was the American Playhouse, which will be part of American Playhouse and starring Jeff Daniels.
Jed Jenkins was a member of American Playhouse, served as Jed Jenkins from 1982 to 1982.
Jed Jenkins was the American Playhouse, a place to play again in 1982 and nominated Jeff Daniels
Jed Jenkins was a play from American Playhouse, serving as Jeff Daniels
Jed Jenkins is a playback that was chosen to be part of American Playhouse in 1982.
Jed Jenkins is an American Playhouse, which will be part of the United States House of Representatives.
Jed Jenkins was a playback from American Playhouse, serving another role in 1982.
Jed Jenkins is a Playhouse which was part of American Playhouse, originally cast as Jeff Daniels.
\end{spverbatim}

\textbf{Input:}
\begin{spverbatim}
[HEAD] It's Great to Be Young (1956 film) [TYPE] starring [TAIL] Cecil Parker [HEAD] Cecil Parker [TYPE] birth year [TAIL] 1897
\end{spverbatim}

\textbf{Expected Outputs:}
\begin{spverbatim}
Cecil Parker, born in 1897, stars in the 1956 film It's Great to be Young.
Cecil Parker, who was born in 1897, starred in the 1956 movie It's Great to Be Young.
It's Great to Be Young starred Cecil Parker, who was born in 1897.\end{spverbatim}

\textbf{Generated Texts:}
\begin{spverbatim}
It's Great to Be Young starred Cecil Parker in both films,
It's Great to Be Young starred as Cecil Parker, who was born in 1897
It's Great to Be Young starred Cecil Parker, in 1897
It's Great to Be Young are Cecil Parker's films, starting in 1897
It's Great to Be Young starred as Cecil Parker, born in 1897
It's Great to Be Young starred Cecil Parker, both in 1897
\end{spverbatim}

\textbf{Input}
\begin{spverbatim}
<page_title> List of ambassadors of the United States to Trinidad and Tobago </page_title> <section_title> List of ambassadors </section_title> <table> <cell> John L. Estrada <col_header> Name </col_header> </cell> </table>
\end{spverbatim}

\textbf{Expected Output}
\begin{spverbatim}
John Learie Estrada is the United States Ambassador to Trinidad and Tobago.
John L. Estrada is the United States Ambassador to Trinidad and Tobago.
John Learie Estrada is the United States Ambassador to Trinidad and Tobago.
\end{spverbatim}

\textbf{Generated texts:}
\begin{spverbatim}
John L. Estrada was an Ambassador from Trinidad and Tobago, which would be part of the United States to Trinidad and Tobago.
John L. Estrada was an ambassador of the United States to Trinidad and Tobago, serving as Trinidad and Tobago.
John L. Estrada was an Ambassador to Trinidad and Tobago, which was chosen to be part of the United States ambassador to Trinidad and Tobago
John L. Estrada was an American Ambassador to Trinidad and Tobago, coming from another country in Trinidad and Tobago.
John L. Estrada was an Ambassador to Trinidad and Tobago, part of the United States delegation to Trinidad and Tobago.
John L. Estrada was an Ambassador to Trinidad and Tobago, which also served as United States Ambassador to Trinidad and Tobago.
John L. Estrada was the United States Ambassador to Trinidad and Tobago, which was served as part of the United States ambassador to Trinidad and Tobago.
John L. Estrada is an American Ambassador to Trinidad and Tobago, which won its place nationwide.
John L. Estrada was an American Ambassador to Trinidad and Tobago, who served as the United States ambassador to Trinidad and Tobago.
John L. Estrada was an Ambassador to Trinidad and Tobago, who served as the United States ambassador to Trinidad and Tobago.
\end{spverbatim}

\textbf{Input:}
\begin{spverbatim}
name[Blue Spice], eatType[coffee shop], customer rating[5 out of 5], near[Crowne Plaza Hotel]

\end{spverbatim}

\textbf{Expected Outputs:}
\begin{spverbatim}
The coffee shop Blue Spice is based near Crowne Plaza Hotel and has a high customer rating of 5 out of 5.
The Blue Spice coffee shop, near Crowne Plaza Hotel, has a customer rating of 5 out of 5.
If you want a coffee shop rated 5 out of 5 pick Blue Spice. It is located near Crowne Plaza Hotel.
\end{spverbatim}

\textbf{Generated Texts:}
\begin{spverbatim}
Blue Spice is a 5 star coffee shop located near Crowne Plaza Hotel that also serves fast food, not family-friendly.
Blue Spice is a 5-star coffee shop located near Crowne Plaza Hotel that also offers food in moderate prices.
Blue Spice is a coffee shop located near Crowne Plaza Hotel that offers great food, at reasonable prices.
Blue Spice is a coffee shop located near Crowne Plaza Hotel that offers cheap food, but only has a 5 out of 5 rating.
Blue Spice is a 5 star coffee shop located near Crowne Plaza Hotel that also serves food in cheap prices.
Blue Spice is a coffee shop near Crowne Plaza Hotel that also offers great food, prices are rated 5 out of 5.
Blue Spice is a 5-star coffee shop located near Crowne Plaza Hotel, that sells sushi for more than £30.
Blue Spice is a coffee shop located near Crowne Plaza Hotel that sells cheap food, and rated 5 out of 5.
Blue Spice is a 5-star coffee shop located near Crowne Plaza Hotel that serves Italian food, and has a price range of more than 30.
Blue Spice is a coffee shop located near Crowne Plaza Hotel that also serves Japanese food, and has a 5 out of 5 rating.
\end{spverbatim}

\end{document}